\documentclass[12pt,arxiv]{l4dc2023}

\title[Black-Box vs. Gray-Box: A Case Study on Table Tennis Ball Trajectory Prediction]{Black-Box vs. Gray-Box: A Case Study on Learning Table Tennis Ball Trajectory Prediction with Spin and Impacts}
\usepackage{times}

\author{%
 \Name{Jan Achterhold} ${}^{1}$ \Email{jan.achterhold@tue.mpg.de}\\
 \Name{Philip Tobuschat} ${}^{2}$ \Email{philip.tobuschat@tue.mpg.de}\\
 \Name{Hao Ma} ${}^{2}$ \Email{hao.ma@tue.mpg.de}\\
 \Name{Dieter Buechler} ${}^{3}$ \Email{dieter.buechler@tue.mpg.de}\\
 \Name{Michael Muehlebach} ${}^{2}$ \Email{michael.muehlebach@tue.mpg.de}\\
 \Name{Joerg Stueckler} ${}^{1}$ \Email{joerg.stueckler@tue.mpg.de}\\
\addr{${}^{1}$ Embodied Vision Group\\${}^{2}$ Learning and Dynamical Systems Group\\
${}^{3}$ Empirical Inference Department\\Max Planck Institute for Intelligent Systems, Tübingen, Germany}
}

\usepackage{bm}
\usepackage{siunitx}
\usepackage{booktabs}
\usepackage{pifont}

\usepackage{wrapfig}
\usepackage{makecell}
\newcommand{\bv}{\bm{v}}

\newcommand{\rx}{\mathrm{x}}
\newcommand{\ry}{\mathrm{y}}
\newcommand{\rz}{\mathrm{z}}
\newcommand{\rai}{\mathrm{ai}}

\newcommand{\rtl}{\mathrm{tl}}
\newcommand{\rtr}{\mathrm{tr}}
\newcommand{\rb}{\mathrm{b}}

\newcommand{\kd}{k_\mathrm{d}}
\newcommand{\km}{k_\mathrm{m}}
\newcommand{\ad}{a_\mathrm{d}}
\newcommand{\am}{a_\mathrm{m}}

\newcommand{\sm}{\bm{s}_\mathrm{m}}

\begin{document}

\maketitle

\begin{abstract}%
In this paper, we present a method for table tennis ball trajectory filtering and prediction. Our gray-box approach builds on a physical model. At the same time, we use data to learn parameters of the dynamics model, of an extended Kalman filter, and of a neural model that infers the ball's initial condition. We demonstrate superior prediction performance of our approach over two black-box approaches, which are not supplied with physical prior knowledge. We demonstrate that initializing the spin from parameters of the ball launcher using a neural network drastically improves long-time prediction performance over estimating the spin purely from measured ball positions. An accurate prediction of the ball trajectory is crucial for successful returns. We therefore evaluate the return performance with a pneumatic artificial muscular robot and achieve a return rate of 29/30 (97.7~\%).
\end{abstract}


\section{Introduction}
Playing table tennis with a robot is a long-standing challenge in robotics research \citep{andersson1989understanding}. An important difficulty resides in tracking and predicting the ball's trajectory, which is strongly influenced by nonlinear effects, arising from drag, spin, and impacts with the table.

In this paper, we focus on designing an accurate model for estimating the ball's state and predicting its future trajectory. Predicting the future trajectory is crucial for computing a hitting point for the robot to return the ball. We build our model on existing knowledge about the physical dynamics of a flying ball, including drag and spin effects. Our method is based on the extended Kalman filter (EKF) and includes various parameters which are trained from offline data.

This gray-box approach demonstrates superior prediction performance compared to two deep-learning based (black-box) baselines in our experiments. In addition, we demonstrate that estimating the ball's initial spin from parameters of the ball launcher with a neural network drastically improves prediction performance over an uninformed initialization of the initial spin. We also highlight the performance of our model by successfully returning balls with a pneumatic artificial muscular robot.

\section{Related Work}
Models for time series can be categorized into white-box, gray-box, and black-box models.
Black-box models follow a purely statistical, data-driven approach without incorporation of prior physical knowledge on the system to model. In contrast, white-box models are purely based on a-priori system knowledge, without incorporation of data. In gray-box models, both physical a-priori knowledge and data are used for model design and the identification of its parameters.
For many dynamical systems, it is difficult or impossible to achieve accurate white-box modeling due to unmodelled effects or variable parameters. Incorporation of data enables black- and gray-box models to adapt to the specifics of the system at hand, which is why we will focus on these two model classes in the following.
Inferring dynamics models and their parameters from data is classically referred to as \emph{system identification}, see \citep{ljung1986system} for an overview.

\paragraph{Black-box models} Black-box time series models often leverage a latent space formulation. They comprise an encoder-decoder pair mapping to and from the latent space, and a forward model in the latent space. Recurrent cells such as long short-term memory (LSTM) \citep{hochreiter1997lstm} or gated recurrent units (GRU) \citep{cho2014properties} are commonly used as latent forward models. Such purely deterministic models can be extended by stochastic nodes to handle noisy observations and transitions, as in variational recurrent neural networks (VRNN) \citep{chung2015recurrent} or stochastic RNNs (SRNN) \citep{fraccaro2016sequential}. Several latent sequence models have been proposed which allow for state estimation through filtering, such as the Kalman-VAE \citep{fraccaro2017disentangled}, Recurrent Kalman Networks \citep{becker2019recurrent}, and Deep Variational Bayes Filters \citep{karl2017deep}. \cite{hafner2019learning} present a recurrent (variational) state-space model (RSSM) which they use for model-based reinforcement learning. We use RSSM as a black-box baseline in our paper.
\citet{girin2021dynamical} present a comprehensive overview of latent sequence models.

\paragraph{Gray-box models} In our work, we follow a gray-box approach, yielding lower prediction error than a black-box baseline and allowing for physical interpretation of the estimated quantities.
In contrast to the black-box models discussed above, gray-box models incorporate prior knowledge about the system to model, such as the laws of physics. One line of work in this direction are \emph{differentiable physics engines} \citep{belbuteperes2018end}. These engines enable gradient-based system identification from data for physical systems with a given structure. Our approach combines ideas from system identification with machine learning. It is therefore capable of exploiting prior knowledge from physics, while also learning parameters of the filter, dynamics, and a state initialization neural network from data.

\subsection{Table tennis ball trajectory modeling}
Approaches for table tennis ball trajectory modelling and prediction can also be categorized as white-box, black-box, and gray-box models. A common \textbf{white-box} approach is to use an aerodynamic model of the ball respecting gravity, drag, and Magnus forces \citep{andersson1989understanding} and a physics-grounded rebound / impact model \citep{nakashima2010modeling} in an extended / unscented Kalman filter \citep{zhang2010_physmodel, mulling2010simulating, wang2014trajectory, zhang2015real, koc2018_optimaltraj, tebbe2018_ekfvspolyfit}. Common \textbf{black-box} models approximate the ball trajectories with polynomial curves which are fitted to recorded data \citep{matsushima2005_llwrmaps, li2012highspeed, tebbe2018_ekfvspolyfit, lin2020_balltracking}.  We compare our approach to the deep-learning based black-box approach by \cite{gomezgonzalez2020real} which leverages a variational auto-encoder architecture for table tennis ball trajectory prediction.
The ball's spin constitutes a particular challenge for table tennis ball trajectory prediction, since it is hard to infer from position measurements of the trajectory. As a result, prior works have resorted to detecting the ball's spin by following the brand logo on the ball \citep{zhang2015real} or by equipping the racket with an inertial sensor \citep{blank2017ball}. In our \textbf{gray-box} approach, we use information from the ball's launch process to initialize the spin. More precisely, we learn the parameters of a neural network that relates the ball launcher settings to the initial spin of the ball. This is shown to drastically improve the quality and accuracy of the predicted ball trajectories.

\section{Method}
We present a gray-box method based on the extended Kalman filter (EKF), which includes parameters that are learned from data.
In our notation, scalars are denoted by lowercase letters ($\sigma$), vectors are denoted by bold lowercase letters ($\bm{\sigma}$), and matrices are denoted by bold uppercase letters ($\bm{\Sigma}$).
The operator $\operatorname{diag}(\bm{x})$ forms a square matrix with $\bm{x}$ on its diagonal. The operator $[x]^+$ applies the softplus function and adds a constant: $[x]^+ = \log (1+e^x) + 10^{-6}$ (elementwise for vectors). 

\subsection{Physical model}
\sloppy We assume the ball's dynamics in free-flight (not impacting with the table) to follow the ordinary differential equation (ODE)
\begin{equation}
\label{eq:ode}
    \dot{\bv}(t) = -k_\mathrm{d} \left\|\bv(t)\right\|_2 \bv(t) + k_\mathrm{m} (\bm{\omega}(t) \times \bv(t)) + \mathbf{g}
\end{equation}
with linear velocity $\bv^\top(t) = (v_\rx(t), v_\ry(t), v_\rz(t))$ and its Euclidean norm $\left\|\bm{v}(t)\right\|_2$, angular velocity (spin) $\bm{\omega}^\top(t) = (\omega_\rx(t), \omega_\ry(t), \omega_\rz(t))$, drag coefficient $\kd$, Magnus effect coefficient $\km$ and gravitational acceleration $\mathbf{g}^\top= \left(0,0,\SI{-9.802}{\metre\per\square\second}\right)$.
We model the table impact by a linear map that relates the pre- and post-impact velocity $\bv$ and spin $\bm{\omega}$ as 
\begin{equation}
\label{eq:impact_model}
((\bv^+)^\top, (\bm{\omega}^+)^\top)^\top = \mathbf{C} ((\bv^-)^\top, (\bm{\omega}^-)^\top)^\top, \quad \mathbf{C} \in \mathbb{R}^{6 \times 6}
\end{equation}
where the superscripts ${}^+$(${}^-$) indicate the linear/angular velocities directly after (before) table impact.

\subsection{Discrete-time state space model}
We formulate a discrete-time state-space model for the ball trajectory for filtering and prediction. 
\paragraph{Free flight}
We introduce a state-space model for the ball dynamics with state 
\begin{equation}
    \bm{z}(t) = (\bm{p}(t)^\top, \bm{v}(t)^\top, \bm{\omega}(t)^\top, \ad(t), \am(t))^\top \in \mathbb{R}^{11}
\end{equation}
where $\bm{p}  \in \mathbb{R}^3$ is the position of the ball's center in Cartesian coordinates. The variables $\ad$, $\am$ parameterize the drag and Magnus coefficients $\kd(t) = \ad^2(t) + \epsilon, \km(t) = \am^2(t) + \epsilon$ to avoid explicit non-negativity constraints and stabilize training. We choose $\epsilon = 0.05$ in our experiments. As in Eq.~\eqref{eq:ode}, $\bv \in \mathbb{R}^3$ and $\bm{\omega}  \in \mathbb{R}^3$ relate to linear and angular velocity, respectively.
To model free-flight phases, we time-discretize the ODE in Eq.~\eqref{eq:ode} by Euler's method
\begin{align}
\label{eq:dyn_discretized}
\begin{split}
    \bm{p}(t+\Delta_t) &= \bm{p}(t) + \Delta_t \bm{v}(t) \\
    \bm{v}(t+\Delta_t) &= \bm{v}(t) + \Delta_t \left(-\kd(t) ||\bm{v}(t)|| \bm{v}(t) + \km(t) (\bm{\omega}(t) \times \bm{v}(t)) + \mathbf{g} \right) \\
    \bm{\omega}(t+\Delta_t) &= \bm{\omega}(t), \quad \ad(t+\Delta_t) = \ad(t), \quad \am(t+\Delta_t) = \am(t)
\end{split}
\end{align}
The function $\bm{z}(t+\Delta_t) = g_\mathrm{free}(\bm{z}(t), \Delta_t)$ abbreviates Eq.~\eqref{eq:dyn_discretized}. To refer to states at discrete time indices $n \in \{1,...,N\}$ we use the notation $\bm{z}_{n+1} = g_\mathrm{free}(\bm{z}_{n}, \Delta_T)$.
In our setting, $\Delta_T = \frac{1}{\SI{180}{\per\second}} \approx \SI{5.56}{\milli\second}$ as the cameras of the video tracking system are triggered with a fixed frequency of $\SI{180}{\per\second}$.

\paragraph{Impact model}
An impact of the ball with the table occurs within a discrete-time increment from $n$ to $n+1$ if the lower edge of the ball (with radius $r$) at $p_{\rz, n+1} - r$ penetrates the table, i.e., $p_{\rz, n+1} - r < \rz_\mathrm{table}$. We approximate the time of impact $\Delta_{\mathrm{imp}} \in [0, \Delta_T]$ with a simplified model to avoid numerical instabilities. It incorporates the velocity of the ball at the last discrete timestep before the impact $v_{\rz,n}$, the gravitational acceleration $g_\rz$, and the height difference to the table $h = -((p_{\rz,n} - r) - \rz_\mathrm{table})$ such that
$
    \Delta_{\mathrm{imp}} = -(v_{\rz,n} + \sqrt{v_{\rz,n} \cdot v_{\rz,n} + 2 g_{\rz} h})/g_{\rz}.
$
The state of the ball just before the impact is given by $\bm{z}^{{-}} = g_\mathrm{free}(\bm{z}_{n}, \Delta_{\mathrm{imp}})$. At the time of impact, the velocity and spin are updated according to Eq.~\eqref{eq:impact_model}, yielding the state $\bm{z}^{{+}}$ after impact. We denote this by $\bm{z}^+=\bm{C}' \bm{z}^-$ with $\bm{C}' = \operatorname{blockdiag}(\bm{I}_3, \bm{C}, 1, 1)$. After the impact a free-flight phase follows, such that at the next discrete timestep, the state is $\bm{z}_{n+1} = g_\mathrm{free}(\bm{z}^{{+}}, \Delta_T - \Delta_{\mathrm{imp}})$.

\paragraph{Joint model}
We denote our discrete-time forward step, incorporating free flight and impacts, as 
\begin{equation}
\label{eq:dyn_joint}
\bm{z}_{n+1} = g(\bm{z}_{n}) =
    \begin{cases}
    g_\mathrm{free}(\bm{z}_{n}, \Delta_T) & \left[ g_\mathrm{free}(\bm{z}_{n}, \Delta_T) \right]_{\rz} - r \geq \rz_\mathrm{table}\\
    g_\mathrm{free}(\mathbf{C}' g_\mathrm{free}(\bm{z}_{n}, \Delta_{\mathrm{imp}}), \Delta_T - \Delta_{\mathrm{imp}}) & \text{otherwise}.
\end{cases}
\end{equation}
where $\left[ \bm{z} \right]_{\rz}$ extracts the $\mathrm{z}$-coordinate of the position in state $\bm{z}$.

\subsection{Extended Kalman Filter (EKF)}
For filtering and prediction, we assume the ball dynamics as in Eq.~\eqref{eq:dyn_joint} with additive Gaussian noise
\begin{equation}
    \hat{\bm{z}}_{n+1} = g(\hat{\bm{z}}_{n}) + \bm{\zeta}, \quad \bm{\zeta} \sim \mathcal{N}(0, \operatorname{diag}([\bm{\sigma}_{{q}}]^+)), \quad \bm{\sigma}_{{q}} \in \mathbb{R}^{11}.
\end{equation}
We obtain measurements of the ball's center $\bm{m}_n \in \mathbb{R}^3$ through a vision tracking system, which we assume to be perturbed by additive Gaussian measurement noise, i.e. $\bm{m}_n = \bm{\hat{p}}_n + \bm{\epsilon}$, $\bm{\epsilon} \sim \mathcal{N}(0, \operatorname{diag}([\bm{\sigma}_{{r}}]^+))$ with $\bm{\sigma}_{{r}} \in \mathbb{R}^{3}$. 
We estimate a belief of the state $p(\bm{\hat{z}}_n \:|\: \bm{m}_{1:n})$, given past position measurements $\bm{m}_{1:n}$, with an extended Kalman filter.  Occasionally, the vision tracking system is unable to compute a position estimate (e.g. due to occlusions), which leads to missing measurements. To this end, we introduce the operator $\tau(n)$, which maps to the index of the $n^\text{th}$ \emph{available} measurement.

\paragraph{State initialization} 
\label{sec:ekf_state_initialization}
We are interested in initializing the state belief at the time when the second measurement is available, i.e. $p(\hat{\bm{z}}_{\tau(2)} \:|\: \bm{m}_{\tau(1)}, \bm{m}_{\tau(2)})$. 
The expected ball's position is estimated by the second measurement, $\bm{p} = \bm{m}_{\tau(2)}$, and the expected velocity $\bm{v}$ by a finite difference approximation $\bm{v} = {(\bm{m}_{\tau(2)}-\bm{m}_{\tau(1)})}/{(\Delta_T(\tau(2) - \tau(1)))}$. 
The initial values for the position and velocity covariance are learned and denoted by $\mathbf{\Sigma}_{{p}} =   \operatorname{diag}([\bm{\sigma}_{{p}}]^+) $, $\mathbf{\Sigma}_{{v}} =  \operatorname{diag}([\bm{\sigma}_{{v}}]^+) $. 
Before a table impact has happened, we can relate the ball's spin to the launcher parameters (as we assume the spin to be constant within the free-flight phase). 
After an impact has happened, this is no longer possible, as the impact changes the initial spin of the ball.
We indicate by $\bm{1}_{\rai}$ whether $\bm{m}_{\tau(2)}$ is taken after an impact. 
To compute a belief for the initial spin, we first assume the launcher's head to be oriented horizontally along the $\rx$-axis. 
For this launch orientation, we compute a ``canonical spin" and its covariance depending on the motor parameters $\sm$ (see Sec.~\ref{sec:launcher}), which we denote by $\bm{\omega}_{\rightarrow \rx} = f^{\omega}(\sm, \boldsymbol{\psi}_f)$, $\mathbf{\Sigma}_{{\omega}_{\rightarrow \rx}} = \operatorname{diag}([f^{\mathbf{\Sigma}_\omega}(\sm, \boldsymbol{\psi}_f)]^+)$. The functions $f^{\omega}, f^{\mathbf{\Sigma}_\omega}$ are implemented by a neural network with two heads with parameters $\boldsymbol{\psi}_f$.
The azimuthal launch orientation can be changed by rotating the whole launcher frame by $\phi_f$ and by rotating the launcher's head by $\phi_l$ (see Fig.~\ref{fig:launcher}(c)).
The elevational launch orientation can be changed by rotating the launcher's head by $\theta_l$. 
To obtain the initial spin in the world coordinate system, we rotate the ``canonical" spin $\bm{\omega}_{\rightarrow \rx}$ accordingly. 
We absorb all rotations in a rotation matrix $\mathbf{R}_\mathrm{rot}(\phi_f + \phi_l, \theta_l)$, such that $\bm{\omega} = \mathbf{R}_\mathrm{rot} \bm{\omega}_{\rightarrow \rx}$, $\mathbf{\Sigma}_\omega = \mathbf{R}_\mathrm{rot} \mathbf{\Sigma}_{{\omega}_{\rightarrow \rx}} \mathbf{R}_\mathrm{rot}^\top$. 
These considerations only hold true for the free-flight phase after ball launch. 
After a table impact has happened, we cannot directly relate the ball spin to the launcher parameters. 
Therefore, in this case, we set the moments of the initial spin $\bm{\omega} = \bm{0}$, $\mathbf{\Sigma}_\omega = \operatorname{diag}([{\bm{\sigma}_{\omega, \rai}}]^+)$. 
During training, we obtain the angles $\phi_l$, $\theta_l$ from piecewise linear regression models which map from launcher parameters $\bm{s}_\phi$, $\bm{s}_\theta$ to $\phi_l$, $\theta_l$. 
We obtained these models on the training split of trajectories recorded from the \emph{default} launcher orientation. 
For the default orientation, we assume the launcher to shoot balls in the $-\ry$ direction, i.e. $\phi_f = -90{}^{\circ}$. 
When evaluating the filter in simulation or on the real robot, we infer the total azimuthal launch angle $\phi_f+\phi_l$ from the first two measurements of the trajectory, to avoid measuring the orientation of the launcher frame $\phi_f$. In summary, we provide the information ($\phi_f$, $\phi_l$, $\theta_l$, $\sm, \bm{1}_{\rai}$) of the ball launch to the model.
As an ablation, we initialize $\bm{\omega} = \bm{0}$, $\mathbf{\Sigma}_\omega = \operatorname{diag}([{\bm{\sigma}_{\omega}}]^+)$, not depending on this information. 
For the drag and Magnus effect coefficient, we learn the mean and covariance of the initial state. The full initial state belief incorporating the first two available measurements is thus given by $\bm{\mu}_{\tau(2)|\tau(1:2)} = (\bm{p}^\top, \bm{v}^\top, \bm{\omega}^\top, {\ad}, {\am})^\top $, $\mathbf{\Sigma}_{\tau(2)|\tau(1:2)} = \operatorname{blockdiag}(\mathbf{\Sigma}_{{p}}, \mathbf{\Sigma}_{v}, \mathbf{\Sigma}_{{\omega}}, [{\sigma_ {\ad}}]^+, [{\sigma_{\am}}]^+)$.
In summary, the parameters of our gray-box model are $\boldsymbol{\Psi} = \left\{ \mathbf{C}, \bm{\sigma}_q, \bm{\sigma}_r, \bm{\sigma}_p, \bm{\sigma}_v, \bm{\psi}_f, \bm{\sigma}_{\omega}, \bm{\sigma}_{\omega,\rai}, \ad, \am, \sigma_{\ad}, \sigma_{\am} \right\}$.

\paragraph{Prediction step} We follow the standard extended Kalman filter prediction step, yielding the prediction mean and covariance matrix $\bm{\mu}_{n+1|1:n} = g(\bm{\mu}_{n|1:n})$, $\mathbf{\Sigma}_{n+1|1:n} = \mathbf{J} \mathbf{\Sigma}_{n|1:n} \mathbf{J}^\top + \mathbf{Q}$, where $\mathbf{J}$ is the Jacobian matrix of the transition model, $\mathbf{J} = \frac{\partial}{\partial\bm{z}_n} g(\bm{z}_n)\vert_{\substack{\bm{z}_n=\bm{\mu}_{n|1:n}}}$.

\paragraph{Correction step}  In the case of missing observations, we perform multiple prediction steps before correcting the state belief with the next available measurement. We now assume that at timestep $n+1$ a measurement is available. For the correction step, we first compute the Kalman gain $\mathbf{K}$ with the observation matrix $\mathbf{H} =  [\mathbf{I}_{3}, \mathbf{0}^{3\times8}]$ as $\mathbf{K} = \mathbf{\Sigma}_{n+1|1:n} \mathbf{H}^\top (\mathbf{H} \mathbf{\Sigma}_{n+1|1:n} \mathbf{H}^\top + \mathbf{R})^{-1}$. From this, we obtain the corrected moments as $\bm{\mu}_{n+1|1:n+1} = \bm{\mu}_{n+1|1:n} + \mathbf{K}(\mathbf{m}_{n+1} - \mathbf{H}\bm{\mu}_{n+1|1:n})$, $\bm{\Sigma}_{n+1|1:n+1} = (\mathbf{I} - \mathbf{K} \mathbf{H}) \bm{\Sigma}_{n+1|1:n}$.

\paragraph{Learning} For notational simplicity, we again assume that there are no missing observations. Let $\mathcal{P} = \{(\hat{\mathbf{m}}^k_1, ..., \hat{\mathbf{m}}^k_{L_k})\}_{k=1}^K$ denote the set of training trajectories, consisting of $K$ sequences of ball position measurements, each sequence being of length $L_k$. The $\operatorname{chunk}$ operator expands a single trajectory into $L+1-N$ chunks of length $N=50$: $
\operatorname{chunk}(\hat{\mathbf{m}}_1, ..., \hat{\mathbf{m}}_{L}) = \{ (\mathbf{m}_i, ..., \mathbf{m}_{i+N-1})\}_{i=1}^{L+1-N}
$.
By $\mathcal{P}_c = \bigcup_{k=1}^K \operatorname{chunk}(\hat{\mathbf{m}}^k_1, ..., \hat{\mathbf{m}}^k_{L_k})$ we denote the set of training chunks and $\Delta(\mathcal{P}_c)$ the distribution over training chunks with uniform probability. For learning the filter parameters~$\boldsymbol{\Psi}$, we maximize their expected marginal log-likelihood under the training chunk distribution,  that is,
$
    \max_{\boldsymbol{\Psi}} \: \mathbb{E}_{\mathbf{m}_{1:N} \sim \Delta(\mathcal{P}_c)} \log p(\mathbf{m}_{3:N} \:|\: \mathbf{m}_1, \mathbf{m}_2, \boldsymbol{\Psi})
$.
As we do not aim to learn a generative model of chunks but are only interested in applying the learned model for filtering and prediction, we additionally condition the marginal log-likelihood on the first two measurements, since we use these for initializing the filter.
The marginal log-likelihood $\log p(\mathbf{m}_{3:N} \:|\: \mathbf{m}_1, \mathbf{m}_2, \boldsymbol{\Psi})$ can be decomposed \citep{sarkka2013bayesian} as follows
\begin{equation}
    \log p(\mathbf{m}_{3:N} \:|\: \mathbf{m}_1, \mathbf{m}_2, \boldsymbol{\Psi}) = \sum_{n=3}^N \log p(\mathbf{m}_n \:|\: \mathbf{m}_{1:n-1}, \boldsymbol{\Psi}) = \sum_{n=3}^N \log \mathcal{N}(\mathbf{m}_n; \bm{\mu}_{n|1:n-1}, \mathbf{\Sigma}_{n|1:n-1})\!
\end{equation}
which allows for an iterative computation with complexity $\mathcal{O}(N)$ when filtering the chunk~$(\mathbf{m}_1,...,\mathbf{m}_N)$. We maximize the expected marginal log-likelihood on batches of chunks with batchsize 64 using the Adam optimizer \citep{kingma2015adam} with learning rate $5\cdot10^{-3}$.

\section{Experiments}
\begin{wrapfigure}{R}{0.25\textwidth}
  \vspace{-2em}
  \begin{center}
    \includegraphics[width=0.23\textwidth]{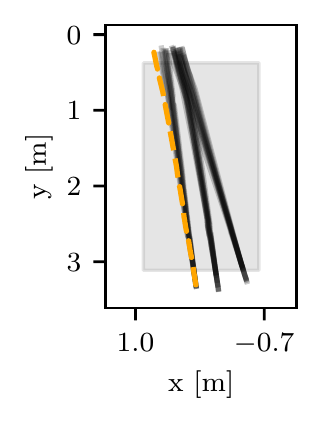}
  \end{center}
  \vspace{-2em}
  \caption[]{\unskip}
  Trajectories of balls launched towards the robot arm (at $\ry \approx 0$), with unreturned trajectories colored orange. 29/30 launches are returned; details: Sec.~\ref{sec:return_experiment}.
  \label{fig:pamy_return}
  \vspace{-3em}
\end{wrapfigure}

We conduct several experiments to answer the following research questions: \textbf{Q1:} How large is the prediction error of the EKF model, and how does it compare to black box baselines? \textbf{Q2:} Does supplying the launch parameters (launch direction, launcher motor speeds) to the predictive model improve prediction performance for the EKF and the RSSM baseline? We use a neural network to infer the initial ball spin from motor parameters, leading us to \textbf{Q3:} Does the spin inferred from the launcher parameters relate to a simple spin model derived from physical principles? Finally, we are interested in the ratio of balls returned by a robot arm \citep{buchler2016lightweight} when using the proposed model for trajectory prediction (\textbf{Q4}).

\subsection{Setup}
\begin{figure}[tb!]
\centering
\subfigure[Setup with the ball launcher (left) and the robot arm (right).]{\includegraphics[width=0.45\textwidth]{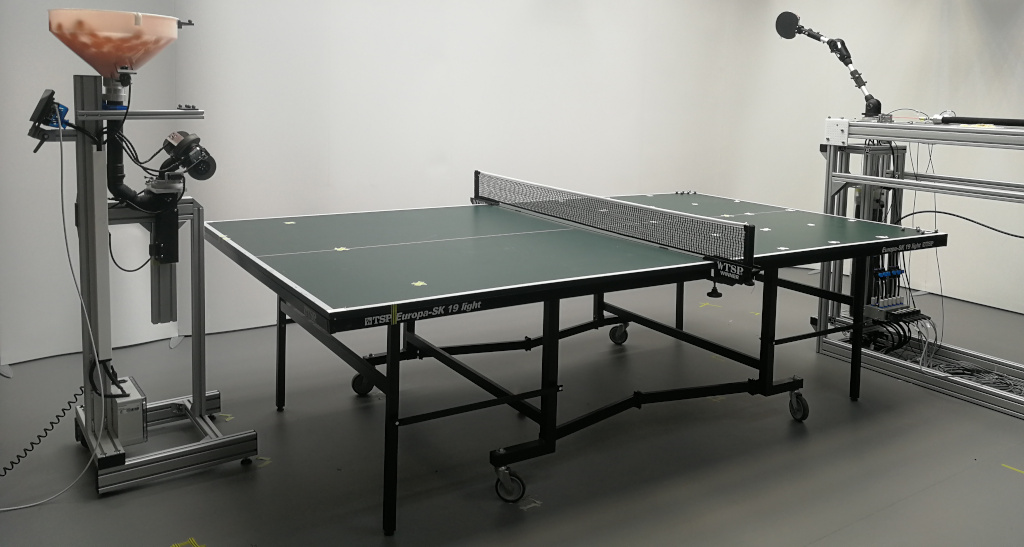}}\hfill
\subfigure[Dataset of recorded table tennis trajectories, launched from the ``default" orientation. Five randomly selected trajectories are colored.]{\includegraphics[width=0.53\textwidth]{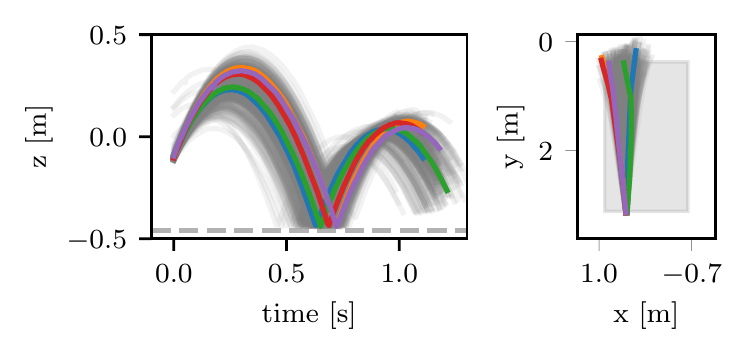}}\hfill
\caption{Experimental setup (a) and visualization of recorded trajectories (b).}
\vspace{-1em}
\label{fig:lab_setup}
\end{figure}

In Fig.~\ref{fig:lab_setup} we show our experimental setup. A ball launcher shoots table tennis balls towards a robot arm that is actuated with pneumatic artificial muscles \citep{buchler2016lightweight}. The position of the ball is measured using four RGB cameras as described in \cite{gomez-gonzalez2019_rt2}. Details of the ball launcher are described in Sec.~\ref{sec:launcher}. The robot is only used for the return experiment in Sec.~\ref{sec:return_experiment}. For training the predictive models, we use recorded trajectories (Sec.~\ref{sec:data_recording}).

\subsubsection{Launcher details}
\label{sec:launcher}
\begin{figure}[tb!]
\centering
\subfigure[Ball launcher]{\includegraphics[height=1.2in]{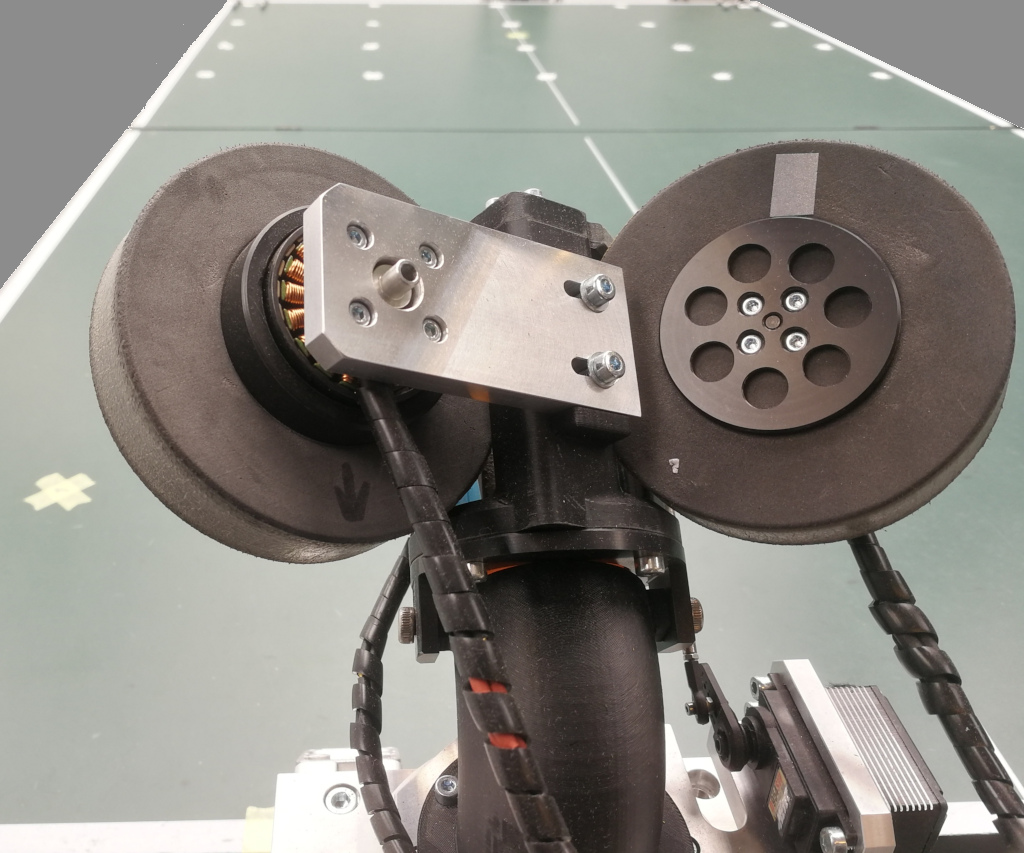}\label{fig:launcher_photo}}\hfill
\subfigure[Schematic drawing (rear view)]{\includegraphics[height=1.2in]{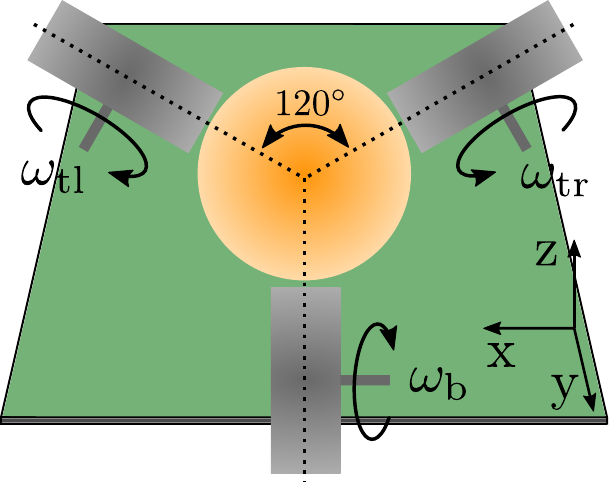}\label{fig:launcher_schematic}}\hfill
\subfigure[Schematic drawing (top/side view) \label{fig:launcher_angles}]{\includegraphics[height=1.2in]{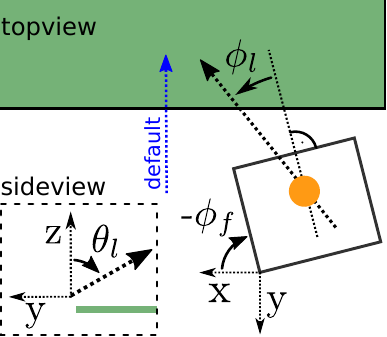}}\hfill
\subfigure[Launch directions]{\includegraphics{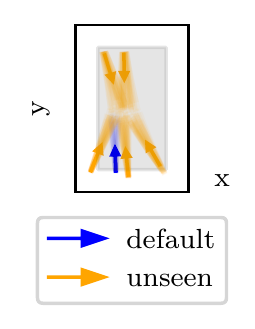}\label{fig:launcher_directions}}
\caption{Experimental setup of the ball launcher. (a) Photo of the launcher (taken in the direction of ball launch, with table in background), (b) a schematic drawing of the launcher with rotating wheels (gray) and ball (orange), (c) frame ($\phi_f$) and launch angles ($\phi_l$, $\theta_l$), (d) shoot directions for the ``default" and ``unseen" configurations.}
\label{fig:launcher}
\vspace{-1em}
\end{figure}

\begin{figure}[tb!]
\centering
\includegraphics[width=\textwidth]{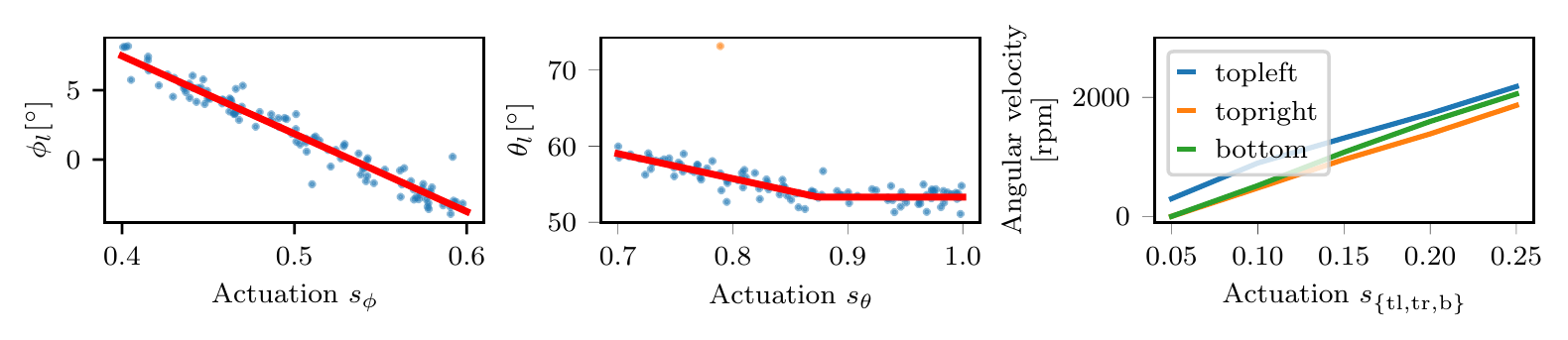}
\vspace{-2em}
\caption{Relation of launcher parameters $\bm{s} \in [0, 1]^5$ to launch angles $\phi_l$, $\theta_l$ and angular velocity of launcher wheels. The left two panels show the result of piecewise linear regression (red) to initial trajectory angles (blue); the right panel shows angular velocity of wheels depending on the actuation. The outlier (orange, center panel) is excluded.}
\vspace{-1em}
\label{fig:launcher_parameters}
\end{figure}

Our ball launcher follows the design of \citet{dittrich2022aimy}. It accelerates the table tennis ball using three rubber wheels which are actuated by brushless motors. The azimuthal and elevational angle of the launcher's head can be adjusted with servo motors to change the direction of the launch. The launcher frame can be freely positioned and rotated about the $\rz$-axis, as parameterized by the angle $\phi_f$. We refer to Fig.~\ref{fig:launcher} for more details on the launcher, including its geometry and the launch angles. The angular velocity of the top-left, top-right, and bottom motor are controlled through three actuation parameters $\sm = (s_{\mathrm{m},\rtl}, s_{\mathrm{m},\rtr}, s_{\mathrm{m},\rb})^\top \in [0, 1]^3$. The mapping from actuation parameters to angular motor velocity is nonlinear, see Fig.~\ref{fig:launcher_parameters}. The azimuthal ($\phi_l$) and elevational ($\theta_l$) launch angles can be controlled through two parameters $s_\phi \in [0, 1], s_\theta \in [0, 1]$. We fit piecewise linear functions to the launch angles of the recorded trajectories to find a mapping from the actuation parameters $s_\phi \in [0, 1], s_\theta \in [0, 1]$ to launch angles $\phi_l$, $\theta_l$ (see Fig.~\ref{fig:launcher_parameters}). The azimuthal launch direction can further be changed by rotating the launcher's frame about the $\rz$-axis by the angle $\phi_f$. For the \emph{default} orientation, we oriented the launcher such that it shoots along the negative $\ry$-axis for $\phi_l = 0$, i.e., $\phi_f = -90{}^{\circ}$. 

\subsubsection{Data recording}
\label{sec:data_recording}
For collecting trajectories for training, validation, and testing, we position the launcher at six different positions and orientations (see Fig.~\ref{fig:launcher}(d)). We term one particular position/orientation \emph{default}, which we use both for training and testing, and the other five \emph{unseen}, which we use for testing only. On the \emph{default} orientation we collect 334 trajectories, which we split in 108 for training, 63 for validation, and 163 for testing. For each of the five \emph{unseen} configurations we collect 30 trajectories which are used for testing only. For each trajectory we randomly sample the launcher parameters uniformly from  $s_\phi \in [0.4, 0.6]$, $s_\theta \in [0.7, 1.0]$, $s_\rtl \in [0.095, 0.155]$ (default), $s_\rtl \in [0.105, 0.165]$ (unseen), $s_{\{\rtr,\rb\}} \in [0.135, 0.195]$ (default), $s_{\{\rtr,\rb\}} \in [0.145, 0.205]$ (unseen). To simulate different launcher orientations, we optionally augment the training data by rotating each trajectory by a random angle $\phi_f \in [0, 2\pi]$ about the $\rz$-axis at the point with minimal $\rz$ coordinate. We add 19 rotated trajectories for each existing trajectory to the training set, which forms the augmented dataset.

\subsection{Prediction performance}
\label{sec:eval_prediction_performance}
\paragraph{Evaluation protocol}
The predictive performance of the investigated models is quantified by measuring the prediction error when filtering until one second before the trajectory ends, and predicting the remaining part of the trajectory. For shorter trajectories, we filter at least ten measurements. The prediction error for each sequence is given by the maximum Euclidean distance between the last five prediction-measurement pairs.

\paragraph{Baselines}
As a first baseline, we train a recurrent state space model, taken from a re-implementation\footnote{\url{https://github.com/Kaixhin/PlaNet}} of \citep{hafner2019learning}. We inherit the standard parameters except for the ”free nats” parameter, which we determined empirically as $0.3$ for minimal average prediction error on the validation split of the \emph{default} dataset. Optionally, we pass the same ball launch information used in the EKF model for state initialization as action to the RSSM model. We train the model for 100,000 steps. As a second baseline, we train a trajectory variational auto-encoder (TVAE) from \cite{gomezgonzalez2020real}, using the provided implementation\footnote{\url{https://github.com/sebasutp/trajectory_forcasting}}. We use a model length of 250 as our longest trajectory is 235 steps. We train the model until the validation loss increases.

\begin{figure}[tb!]
\centering
\subfigure[All methods (change of scale from linear to log at $y=1$)]{\includegraphics{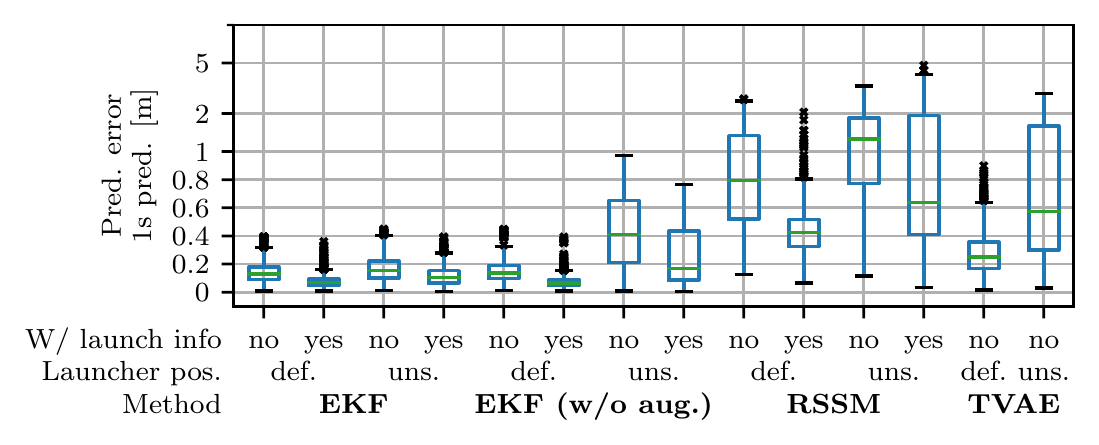}}\hfill
\subfigure[Detailed view for EKF]{\includegraphics{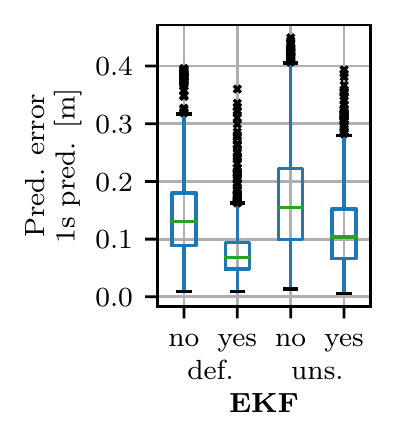}}
\caption{Prediction error for various methods on the test set of default (def.) and unseen (uns.) launcher positions, with and without ball launch information (launch info), for a prediction horizon of one second (see Sec.~\ref{sec:eval_prediction_performance}). Depicted statistics are over the prediction errors for ten independently trained models. All models are trained on the augmented dataset, except ``EKF w/o aug.". The EKF model outperforms the RSSM \citep{hafner2019learning} and TVAE \citep{gomezgonzalez2020real} models (a). The EKF's prediction error can further be reduced by providing ball launch information (b).}
\vspace{-1em}
\label{fig:results}
\end{figure}

\begin{figure}[tb!]
\centering
\includegraphics[width=0.33\textwidth]{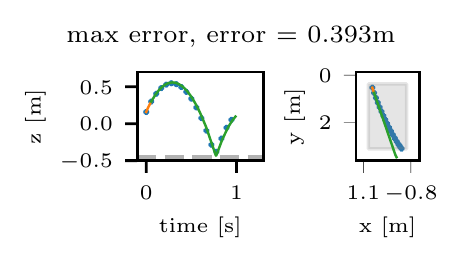}\hfill
\includegraphics[width=0.33\textwidth]{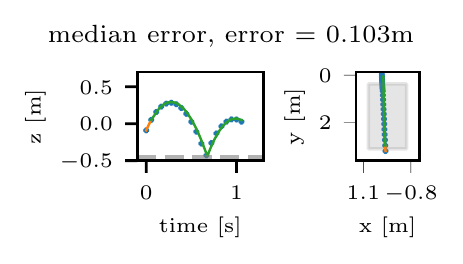}\hfill
\includegraphics[width=0.33\textwidth]{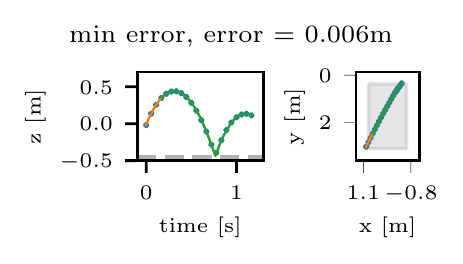}
\vspace{-1em}
\caption{EKF filtering / prediction results on the \emph{unseen} launcher orientations (with ball launch information). We filter until one second before the end of the trajectory (orange) and predict the remaining one second (green). Measurements are colored blue. We show every tenth measurement for visual clarity. Shown are the trajectories with max. / median / min. prediction error at the end of the trajectory from all \emph{unseen} launcher configurations, over ten independently trained models. The median prediction error is \SI{10.3}{\centi\metre}.}
\label{fig:filter_trajectories}
\vspace{-1em}
\end{figure}

\paragraph{Results}
We refer to Fig.~\ref{fig:results} for a visualization of the results. We observe that augmenting the training data is important for the EKF approach presented herein to generalize to the \emph{unseen} launcher positions (Fig.~\ref{fig:results}(a)). In all settings, the EKF approach shows superior performance compared to the RSSM and TVAE baselines. Fig.~\ref{fig:results}(b) shows that initializing the spin using ball launch information reduced the prediction error drastically, both on the \emph{default} and \emph{unseen} launcher configurations. We show representative filtering and prediction results on three trajectories in Fig.~\ref{fig:filter_trajectories}.

\subsection{Spin evaluation}
\label{sec:spin_eval}
\begin{figure}[tb!]
    \centering
    \includegraphics[width=\textwidth]{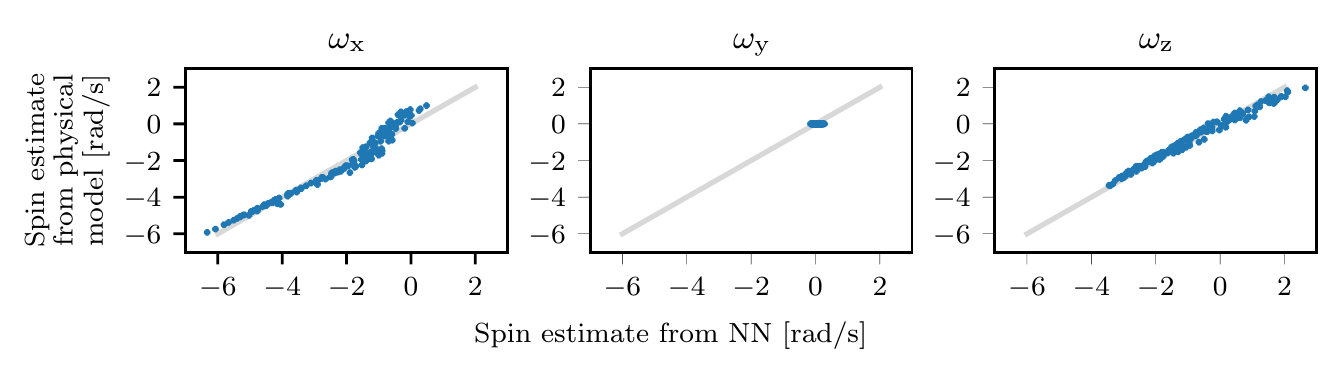}
    \vspace{-2em}
    \caption{Correlation of spin values estimated by the neural network (NN) $f^\omega(\sm, \bm{\psi}_f)$ from motor actuations $\sm$ and spin values computed with Eq.~\eqref{eq:spin_model}. Both estimated spin values highly correlate, indicating physical plausibility of spins estimated by $f^\omega(\sm, \bm{\psi}_f)$, see Sec.~\ref{sec:spin_eval} for details.}
    \label{fig:spin_evaluation_correlation}
    \vspace{-1em}
\end{figure}

With this experiment, we aim to verify the plausibility of spins which are estimated by the learned neural network $f^\omega(\sm,\bm{\psi}_f)$ given launcher parameters $\sm$ (see Sec.~\ref{sec:ekf_state_initialization}). For this, we formulate a simple model for the ball spin, which is derived from the geometry of the launcher (see Fig.~\ref{fig:launcher}(b)). We model the ball's spin for a launcher which is oriented to shoot balls in the $-\ry$ direction as
\begin{align}
    \label{eq:spin_model}
    \left(
        \omega_{\rx, \overrightarrow{-y}},
        \omega_{\ry, \overrightarrow{-y}},
        \omega_{\rz, \overrightarrow{-y}}
    \right)^\top =
    \omega_{\rtl} \alpha \left(
        \frac{1}{2},
        0,
        -\frac{\sqrt{3}}{2}
    \right)^\top +
    \omega_{\rtr} \beta \left(
        \frac{1}{2},
        0,
        \frac{\sqrt{3}}{2}
    \right)^\top +
    \omega_{\rb} \gamma
    \left(
        -1,
        0,
        0
    \right)^\top.
\end{align}
The reasoning behind the model is that every motor adds a spin component to the ball, which is the motor speed ($\omega_{\rtl}, \omega_{\rtr}, \omega_{\rb}$) scaled by a constant ($\alpha, \beta, \gamma$). We note that the bottom motor causes a spin in negative $\ry$ direction. The direction of the spin of the top motors is obtained by rotating the bottom-motor spin unit vector $(-1, 0, 0)^\top$ by $120^{\circ}$ ($240^{\circ}$) about the $\ry$-axis. For this experiment, we obtain the values for $\omega_{\rtl,\rtr,\rb}$ from measurements for the angular launcher wheel velocity given the actuation parameters (see Fig.~\ref{fig:launcher_parameters}(c)). For all \emph{test} trajectories from the \emph{default} dataset, we first compute the spin for a launch in $\rx$ direction with $f^\omega(\sm,\bm{\psi}_f)$. We rotate this spin by $-90^{\circ}$ about the $\rz$-axis to obtain the spin in $-\ry$ direction, as in Eq.~\eqref{eq:spin_model}. Finally, we obtain the parameters $\alpha,\beta,\gamma$ by minimizing a squared error between the rotated spins from $f^\omega$ and the spins estimated by Eq.~\eqref{eq:spin_model}. In Fig.~\ref{fig:spin_evaluation_correlation} we show that the two spin estimates highly correlate, indicating that the values $f^\omega(\sm,\bm{\psi}_f)$ indeed relate to the actual spin of the ball. It is important to note that without additional physical information, we can determine the spin only up to a scaling factor. This is because we both learn $\km = \am^2 + \epsilon$ and $\bm{\omega}$ which appear as a product in Eq.~\eqref{eq:ode}. The actual spin could be inferred as $\bm{\omega}^* = \km \bm{\omega} / {\km^*}$ with $\km^* = C_\mathrm{m} \rho A r / (2m)$ for known values of the ball's mass $m$, Magnus lift coefficient $C_\mathrm{m}$, ball radius $r$, air density $\rho$ and cross-sectional area $A=\pi r^2$.

\subsection{Return performance}
\label{sec:return_experiment}
We evaluate the performance of our ball motion predictions by intercepting and returning balls with a four-degrees-of-freedom robot arm, where each degree of freedom is controlled by a pair of pneumatic artificial muscles (PAMs) \citep{buchler2016lightweight, buechler2023learning}.
The robot arm is controlled by a learning-based iterative control framework for trajectory tracking~\citep{ma2022learning}.
A table tennis racket is attached to the robot arm in order to return balls.
As the ping-pong ball flies through the air, its position, velocity, and spin are continuously estimated with the EKF presented herein. 
The future evolution of the ball's states is then simulated in a receding horizon scheme using the learned model of the ball dynamics, with the latest state estimate as the initial condition. 
This predicted ball trajectory is used to determine the interception of the ball with the racket, which we represent as a pair of position (the interception position of the ball and the racket) and time (the time of interception). 
The predicted trajectory, and therefore the interception point, is repeatedly recalculated, as the state estimate is updated and improved with new measurements of the ball. 
The robot arm successfully returns 29 of 30 (97.7~\%) launched balls, leveraging the EKF for filtering and prediction presented above. We refer to Fig.~\ref{fig:pamy_return} for a visualization of returned and unreturned trajectories.

\section{Conclusion}
Based on a physically grounded model for the aerodynamic behavior of a flying ball respecting Magnus and drag effects and the extended Kalman filter, we have designed a filter and predictive model for table tennis ball trajectories. As we fit the parameters of the filter on offline data, no tedious tuning of initial, transition, and observation covariances is required. Our formulation allows for learning a neural model which estimates the ball's initial spin from ball launch information. This drastically improves the performance of long-term predictions compared to an uninformed initialization. 
Our results also support the findings of other works, which state that the ball's spin can only insufficiently be estimated from ball position measurements alone \citep{zhang2015real, blank2017ball}. 
Our method could constitute groundwork for future research which, e.g., incorporates information on the racket movement to estimate the ball's spin and predict its future trajectory with high accuracy.

\section{Acknowledgements}
The authors thank Bernhard Schoelkopf, director of the Empirical Inference Department at the Max Planck Institute for Intelligent Systems, for providing access to the table tennis experiment setup and Alexander Dittrich for providing measurements on the relation of launcher motor actuation parameters and angular velocities. 
The authors thank the International Max Planck Research School for Intelligent Systems (IMPRS-IS) for supporting Jan Achterhold. Hao Ma thanks the Center for Learning Systems (CLS) for the generous support. Michael Muehlebach thanks the German Research Foundation and the Branco Weiss Fellowship, administered by ETH Zurich, for the generous support.

\bibliography{literature}

\end{document}


\maketitle

In the following, we provide extended results and architectural details of our table tennis ball trajectory filtering and prediction approach. In Section~\ref{sec:ttbm_suppl_varhorz} we show prediction errors for varying prediction horizons. In Section~\ref{sec:ttbm_suppl_return} we visualize trajectories of balls which the artificial muscular robot failed to return, for the model variant which is not provided with ball launch information. In section~\ref{sec:ttmb_suppl_arch} we give details on the architecture, in particular the initial and learned values of the EKF parameters, and the network for estimating the initial spin. Section~\ref{sec:ttbm_suppl_jac} describes the computation of the Jacobian of the forward model $\bm{J} = \frac{\partial}{\partial\bm{z}} g(\bm{z})$ (cf. Equation~5 in the main paper).

\section{Prediction error for varying prediction horizons}
\label{sec:ttbm_suppl_varhorz}
\begin{figure}
\centering
\subfigure[All methods.]{\includegraphics[width=0.48\textwidth]{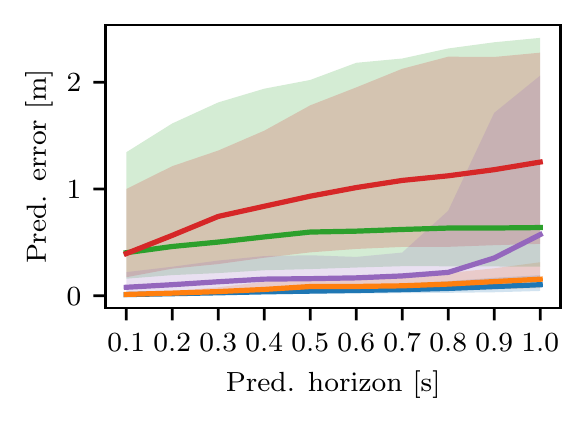}}\hfill
\subfigure[EKF methods only (note different scaling of $y$-axis).]{\includegraphics[width=0.48\textwidth]{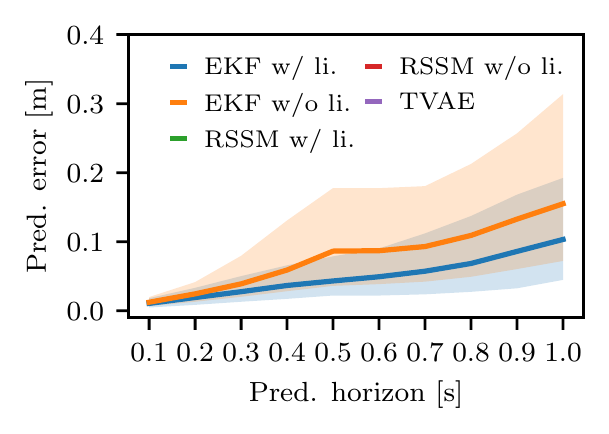}}\hfill
\caption{Prediction error for varying prediction horizons. We evaluated the prediction error on 150 evaluation trajectories from the \emph{unseen} dataset, for 10 independently trained models per method. We show the median error as a solid line, the shaded area covers values between the $10^{\mathrm{th}}$ and $90^{\mathrm{th}}$ percentile. We compare variants with (w/ li.) and without (w/o li.) providing launch information.}
\label{fig:ttbm_suppl_varhorz}
\end{figure}

We refer to Figure~\ref{fig:ttbm_suppl_varhorz} for a visualization of the prediction error for varying prediction horizons. We filter the trajectory from its beginning until the respective prediction horizon remains. For each trajectory, the prediction error is given by the maximum Euclidean distance over the last five measurement-prediction pairs, as stated in the main paper. The proposed EKF method clearly outperforms the baselines RSSM \citep{hafner2019learning} and TVAE \citep{gomezgonzalez2020real} for all considered horizons. Supplying ball launch information (w/ li.) generally reduces the prediction error.

\section{Return experiment}
\label{sec:ttbm_suppl_return}
\begin{figure}
\centering
\subfigure[Launch information is provided to the EKF: 29/30 launches are returned.]{\includegraphics[width=0.3\textwidth]{graphics/return_evaluation.pdf}}\hspace{1em}
\subfigure[Launch information is \emph{not} provided to the EKF: 26/30 launches are returned.]{\includegraphics[width=0.3\textwidth]{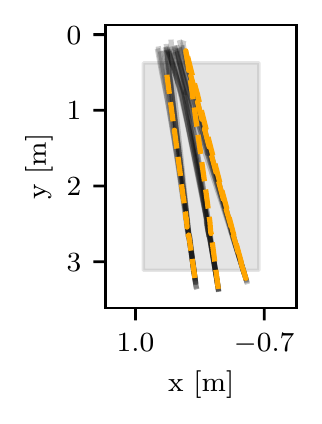}}\hfill
\caption{Trajectories of balls launched towards the robot arm (at $\ry \approx 0$), with unreturned trajectories colored orange.}
\label{fig:ttbm_suppl_return}
\end{figure}

In Figure~\ref{fig:ttbm_suppl_return} we show trajectories of balls launched towards the robot arm for returning, highlighting those which were not returned successfully. In addition to the visualization from the main paper (Fig.~\ref{fig:ttbm_suppl_return}a, launch information is provided, 29/30 trajectories returned), we show failed return trajectories for a setting where no launch information was given to the extended Kalman filter for initializing the ball's spin (Fig.~\ref{fig:ttbm_suppl_return}b). In this case, 26/30 balls were returned.

\section{Architectural details}
\label{sec:ttmb_suppl_arch}
\subsection{Parameter initialization}
In this section, we provide initial values for the parameters
\begin{equation*}
    \boldsymbol{\Psi} = \left\{ \mathbf{C}, \bm{\sigma}_q, \bm{\sigma}_r, \bm{\sigma}_p, \bm{\sigma}_v, \bm{\psi}_f, \bm{\sigma}_{\omega}, \bm{\sigma}_{\omega,\rai}, \ad, \am, \sigma_{\ad}, \sigma_{\am} \right\}
\end{equation*} we optimize during training.

We initialize the impact matrix $\bm{C}$ with the assumption that the velocities in $\rx$- and $\ry$-direction do not change during impact, while the velocity in $\rz$ direction flips sign. We assume the spin to stay constant in all dimensions. This yields
\begin{equation*}
    \bm{C} = \begin{bmatrix}
    1 & 0 & 0 & 0 & 0 & 0 \\
    0 & 1 & 0 & 0 & 0 & 0 \\
    0 & 0 & -1 & 0 & 0 & 0 \\
    0 & 0 & 0 & 1 & 0 & 0 \\
    0 & 0 & 0 & 0 & 1 & 0 \\
    0 & 0 & 0 & 0 & 0 & 1 \\
    \end{bmatrix}
\end{equation*} as initial value for $\bm{C}$. The parameter $\bm{\psi}_f$ refers to the parameters of the spin initialization network, which is detailed in Sec.~\ref{sec:ttbm_suppl_spininit}. Initial values for the remaining parameters are given in Table~\ref{tab:ttbm_suppl_init}.

\begin{table}[h!]
\centering
\begin{tabular}{rlrl}
\toprule
Parameter name & Initial value & Parameter name & Initial value \\
\midrule
Initial state variance & & Transition variance & \\
     $\bm{\sigma}_p$ & $\operatorname{inv}_{[\cdot]^{+}}(10^{-4}) \cdot \vone_3$ & $\bm{\sigma}_q[1:3]$ & $\operatorname{inv}_{[\cdot]^{+}}(10^{-4}) \cdot \vone_3$  \\
     $\bm{\sigma}_v$ & $\operatorname{inv}_{[\cdot]^{+}}(10^{-2}) \cdot \vone_3$  & $\bm{\sigma}_q[4:6]$ & $\operatorname{inv}_{[\cdot]^{+}}(10^{-2}) \cdot \vone_3$  \\
     $\bm{\sigma}_{\omega}, \bm{\sigma}_{\omega,\rai}$ & $\operatorname{inv}_{[\cdot]^{+}}(1) \cdot \vone_3$ &  $\bm{\sigma}_q[7:9]$ & $\operatorname{inv}_{[\cdot]^{+}}(10^{-3}) \cdot \vone_3$  \\
     $\bm{\sigma}_{a_\mathrm{d}}$ & $\operatorname{inv}_{[\cdot]^{+}}(10^{-2})$  & $\bm{\sigma}_q[10]$ & $\operatorname{inv}_{[\cdot]^{+}}(10^{-2})$  \\
     $\bm{\sigma}_{a_\mathrm{m}}$ & $\operatorname{inv}_{[\cdot]^{+}}(10^{-2})$ & $\bm{\sigma}_q[11]$ & $\operatorname{inv}_{[\cdot]^{+}}(10^{-2})$  \\
     $\bm{\sigma}_r$ & $\operatorname{inv}_{[\cdot]^{+}}(10^{-3})  \cdot \vone_3$  \\
     Initial state mean & & & \\
     $a_\mathrm{d}$ & $\sqrt{0.1}$  \\
     $a_\mathrm{m}$ & $\sqrt{0.1}$  \\
\bottomrule
\end{tabular}
\caption{Initial values for parameters of the extended Kalman filter. The operator $\operatorname{inv}_{[\cdot]^{+}}(x)$ inverts the softplus function $[\cdot]^+$, such that, e.g., the initial value for the initial position covariance is $\Sigma_p = \operatorname{diag}([\bm{\sigma}_p]^+) = \bm{I}_{3} \cdot 10^{-4}$. The operator $[\cdot:\cdot]$ denotes 1-based slicing, including start and end point. E.g., $\bm{\sigma}_q[1:3]$ refers to the transition variance of the ball's position. Note that $\bm{\sigma}_{\omega}, \bm{\sigma}_{\omega,\rai}$ are only used when the spin initialization network is not used or after an impact has happened. In these cases, we assume the spin to highly vary, which is why we chose the values of $\bm{\sigma}_{\omega}, \bm{\sigma}_{\omega,\rai}$ to be comparably large.}
\label{tab:ttbm_suppl_init}
\end{table}

\subsection{Learned parameters}
In Table~\ref{tab:ttbm_suppl_learned} we provide the values of $\boldsymbol{\Psi}$ after training for an exemplary EKF model (trained on the augmented dataset, with providing ball launch information). For the impact matrix $\bm{C}$ we learn
\begin{equation}
    \bm{C} = \left[
\begin{array}{c|c}
\bm{C}_\mathrm{vv} & \bm{C}_\mathrm{v\omega} \\
\hline
\bm{C}_\mathrm{\omega v} & \bm{C}_\mathrm{\omega \omega}
\end{array}
\right] =
\left[
\begin{array}{c|c}
\begin{matrix}
\mathbf{0.54}  &  -0.01 &  -0.00 \\
 0.01 &  \mathbf{0.55}  &  0.00 \\
-0.01 &  -0.00 &  \mathbf{-0.92} \\
\end{matrix} &
\begin{matrix}
-0.01 &  \mathbf{0.12}  &  0.01 \\
\mathbf{-0.11}  &  -0.01 &  -0.01 \\
0.01 &  0.00 &  0.00
\end{matrix} \\
\hline
\begin{matrix}
0.19  &  \mathbf{-1.40}  &  0.01\\
\mathbf{1.44}  &  0.14  &  -0.03\\
0.01 &  -0.05 &  \mathbf{0.15}
\end{matrix} &
\begin{matrix}
\mathbf{-0.20}  &  0.09 &  -0.03 \\
-0.01 &  \mathbf{-0.12}  &  -0.01 \\
0.05 & -0.05 &  \mathbf{1.37}
\end{matrix}
\end{array}
\right].
\end{equation}
We interpret $\bm{C}$ to be composed of four submatrices, mapping velocities and spins before to velocities and spins after impact. For each row in each submatrix, we highlight the dominating component. Submatrix $\bm{C}_\mathrm{vv}$ maps velocities before impact to velocities after impact. Velocities are dampened in all directions $\rx, \ry, \rz$ and have no interacting effects. As expected, only the velocity in $\rz$-direction flips sign. Submatrix $\bm{C}_\mathrm{v\omega}$ maps spins before impact to velocities after impact. A positive spin about the $\ry$-axis increases the $\rx$-velocity after impact ($\omega_{\ry}{\mkern-10mu}\uparrow\,\rightarrow\!v_{\rx}{\mkern-10mu}\uparrow$); a positive spin about the $\rx$-axis decreases the $\ry$-velocity after impact ($\omega_{\rx}{\mkern-6mu}\uparrow\,\rightarrow\!v_{\ry}{\mkern-6mu}\downarrow$). Investigating $\bm{C}_\mathrm{\omega v}$, analogously, a negative $\ry$-velocity increases spin about the $\rx$-axis ($v_{\ry}{\mkern-6mu}\downarrow\,\rightarrow\!\omega_{\rx}{\mkern-6mu}\uparrow$), and a positive $\rx$-velocity increases spin about the $\ry$-axis ($v_{\rx}{\mkern-6mu}\uparrow\,\rightarrow\!\omega_{\ry}{\mkern-6mu}\uparrow$). A negative velocity in $z$ direction (i.e., towards the table) reduces the spin about $\rz$. These relations are physically plausible, considering the geometry of the table setup. That our simplified linear model only approximately captures physical reality becomes apparent when investigating the bottom-right entry of $\bm{C}$, which suggests that the spin about the $\rz$-axis is amplified by the impact. We hypothesize that this effect cancels with the spin attenuation due to negative $\rz$-velocities. We leave posing constraints on $\bm{C}$, e.g. for rotational symmetries, and nonlinear impact effects, for future work.

\begin{table}[h!]
\centering
\begin{tabular}{rlrl}
\toprule
Parameter name & Initial value & Parameter name & Initial value \\
\midrule
Initial state variance & & Transition variance & \\
     $[\bm{\sigma}_p]^+$ & $(1.38, 1.63, 1.00)^\top \cdot 10^{-6}$ & $[\bm{\sigma}_q[1:3]]^+$ & $(1.00, 1.00, 1.00)^\top \cdot 10^{-6}$  \\
     $[\bm{\sigma}_v]^+$ & $(0.12, 0.14, 0.09)^\top$  & $[\bm{\sigma}_q[4:6]]^+$ & $(1.16, 1.19, 1.10)^\top \cdot 10^{-6}$  \\
     $[\bm{\sigma}_{\omega,\rai}]^+$ & $(0.19, 3.28, 0.12)^\top$ &  $[\bm{\sigma}_q[7:9]]^+$ & \makecell{$\begin{pmatrix} 1.73 \cdot 10^{-3} \\   1.31 \cdot 10^{-3} \\ 1.00 \cdot 10^{-6} \end{pmatrix}$}  \\
     $[\bm{\sigma}_{a_\mathrm{d}}]^+$ & $5.70 \cdot 10^{-6}$  & $[\bm{\sigma}_q[10]]^+$ & $1.16 \cdot 10^{-6}$  \\
     $[\bm{\sigma}_{a_\mathrm{m}}]^+$ & $3.10 \cdot 10^{-3}$ & $[\bm{\sigma}_q[11]]^+$ & $1.30 \cdot 10^{-3}$  \\
     $[\bm{\sigma}_r]^+$ & $(1.02, 1.03, 1.00)^\top \cdot 10^{-6}$  \\
     Initial state mean & & & \\
     $a_\mathrm{d}$ & ${0.2168}$  \\
     $a_\mathrm{m}$ & ${1.22 \cdot 10^{-5}}$  \\
\bottomrule
\end{tabular}
\caption{Learned values for parameters of the extended Kalman filter, for a model which leverages launch information (thus, $\bm{\sigma}_{\omega}$ is not used).}
\label{tab:ttbm_suppl_learned}
\end{table}

\subsection{Spin initialization network architecture}
\label{sec:ttbm_suppl_spininit}
The neural network for initializing the initial spin in canonical launch direction $\bm{w}_{\rightarrow x}$ based on actuation parameters $\bm{s}_m \in [0, 1]^3$ is parametrized as follows
\begin{equation*}
    \bm{w}_{\rightarrow x, \mu,\sigma} = W_2 \operatorname{max} (W_1 \bm{s}_m, 0),
\end{equation*}
with $\bm{w}_{\rightarrow x, \mu,\sigma} \in \mathbb{R}^{6}$, $W_1 \in \mathbb{R}^{256 \times 3}$, $W_2 \in \mathbb{R}^{6 \times 256}$. The $\operatorname{max}$ operator is applied elementwise. The initial mean $\bm{w}_{\rightarrow x}$ is taken as the first three dimensions of $\bm{w}_{\rightarrow x, \mu,\sigma}[1:3]$, the latter three dimensions parametrize the initial covariance as $\bm{\Sigma}_{\omega \rightarrow x} = \operatorname{diag}([\bm{w}_{\rightarrow x, \mu,\sigma}[4:6]]^+)$.

\section{Derivatives}
\label{sec:ttbm_suppl_jac}
For the extended Kalman filter, we need to compute derivatives (Jacobians) of the forward dynamics $\mathbf{J} = \frac{\partial}{\partial\bm{z}_n} g(\bm{z}_n)$. For computational considerations, we manually implemented the derivatives. The dynamics depend on whether an impact has happened. We restate Equation~5 from the main paper:
\begin{equation*}
\label{eq:dyn_joint}
\bm{z}_{n+1} = g(\bm{z}_{n}) =
    \begin{cases}
    g_\mathrm{free}(\bm{z}_{n}, \Delta_T) & \left[ g_\mathrm{free}(\bm{z}_{n}, \Delta_T) \right]_{\rz} - r \geq \rz_\mathrm{table}\\
    g_\mathrm{free}(h(g_\mathrm{free}(\bm{z}_{n}, \Delta_{\mathrm{imp}})), \Delta_T - \Delta_{\mathrm{imp}}) & \text{otherwise}.
\end{cases}
\end{equation*}
The function $h$ maps the state prior to the impact $\bm{z}^- = g_\mathrm{free}(\bm{z}_{n}, \Delta_{\mathrm{imp}})$ to the state after the impact $\bm{z}^+ = h(\bm{z}^-)$. We abbreviate the remaining time after the impact as $\Delta_{\mathrm{rem}} = \Delta_T - \Delta_{\mathrm{imp}}$. First, we consider free-flight dynamics, and compute the Jacobian
\begin{equation}
    \label{eq:jac_z}
    \bm{J}_{\bm{z}}(\bm{z}_n, \Delta_t) = \frac{\partial}{\partial\bm{z}'_n} g_\mathrm{free}(\bm{z}'_n, \Delta_t) \vert_{\substack{\bm{z}'_n=\bm{z}_n}}
\end{equation}
for an arbitrary timespan $\Delta_t$ given by the components \\
\begin{center}
\begin{tabular}{rccccc}
     & $\cdot/\partial \bm{p}_n$ & $\cdot/\partial \bm{v}_n$ & $\cdot/\partial \bm{\omega}_n$ & $\cdot/\partial a_{\rd,n}$ & $\cdot/\mathrm{d}a_{\rrm,n}$ \\
     $\partial \bm{p}_{n+1}/\cdot$ & $\bm{I}$ & $\Delta_t \bm{I}$ & 0 & 0 & 0 \\
     $\partial \bm{v}_{n+1}/\cdot$ & 0 & Eq.~\ref{eq:dvv} & Eq.~\ref{eq:dvw} & Eq.~\ref{eq:dvad} & Eq.~\ref{eq:dvam} \\
     $\partial \bm{\omega}_{n+1}/\cdot$ & 0 & 0 & $\bm{I}$ & 0 & 0 \\
     $\partial a_{\rd,n+1}/\cdot$ & 0 & 0 & 0 & 1 & 0 \\
     $\partial a_{\rrm,n+1}/\cdot$ & 0 & 0 & 0 & 0 & 1 \\
\end{tabular}
\end{center}
with the following velocity derivatives
\begin{align}
    \label{eq:dvv}
    \frac{\partial \bm{v}_{n+1}}{\partial \bm{v}_n} &= \bm{I} - \Delta_t  a_{\rd,n}^2 \frac{\partial (||\bm{v}_{n}||\bm{v}_{n})}{\partial \bm{v}_n} + \Delta_t  a_{\rrm,n}^2 \frac{\partial (\bm{\omega}_{n} \times \bm{v}_{n})}{\partial \bm{v}_n} \\
    \label{eq:dvw}
    \frac{\partial \bm{v}_{n+1}}{\partial \bm{\omega}_n} &= \Delta_t  a_{\rrm,n}^2 \frac{\partial (\bm{\omega}_{n} \times \bm{v}_{n})}{\partial \bm{\omega}_n} \\
    \label{eq:dvad}
    \frac{\partial \bm{v}_{n+1}}{\partial a_{\rd,n}} &= -2 \Delta_t a_{\rd,n} ||\bm{v}_n|| \bm{v}_n \\
    \label{eq:dvam}
    \frac{\partial \bm{v}_{n+1}}{\partial a_{\rrm,n}} &= 2 \Delta_t a_{\rrm,n} (\bm{\omega}_n \times \bm{v}_n).
\end{align}
Let us now consider the case where an impact has happened. First, it is important to note that the impact time $\Delta_\mathrm{imp}$ (and thus also the remaining time $\Delta_\mathrm{rem}$) depend on $\bm{z}_n$, in particular $v_{\rz,n}$, as
\begin{equation*}
    \Delta_{\mathrm{imp}} = -\left(v_{\rz,n} + \sqrt{v_{\rz,n}^2 + 2 g_{\rz} h}\right)/g_{\rz}.
\end{equation*}
We can thus write
\begin{equation*}
    \bm{z}_{n+1} = g_\mathrm{free}(h(g_\mathrm{free}(\bm{z}_{n}, \Delta_{\mathrm{imp}}(\bm{z}_n)), \Delta_{\mathrm{rem}}(\bm{z}_n))
\end{equation*}
and obtain
\begin{equation*}
    \frac{\partial \bm{z}_{n+1}}{\partial \bm{z}_n} =  \underbrace{\frac{\partial g_\mathrm{free}(\bm{z}^+, \Delta_\mathrm{rem})}{\partial \bm{z}^+}}_{J1} \underbrace{\frac{\partial \bm{z}^+}{\partial \bm{z}_n}}_{J2} + \underbrace{\frac{\partial g_\mathrm{free}(\bm{z}^+, \Delta_\mathrm{rem})}{\partial \Delta_\mathrm{rem}}}_{J3} \underbrace{\frac{\partial \Delta_\mathrm{rem}}{\partial \bm{z}_n}}_{J4}.
\end{equation*}
with $\bm{z}^+ = h(g_\mathrm{free}(\bm{z}_{n}, \Delta_{\mathrm{imp}}(\bm{z}_n))$.
The term $(J1) = \bm{J}_{\bm{z}}(\bm{z}^+, \Delta_\mathrm{rem})$ is given by the Jacobian in Eq.~\ref{eq:jac_z}. For later use, we introduce the Jacobian
\begin{equation}
    \label{eq:jac_t}
    \bm{J}_{\Delta_t}(\bm{z}_n, \Delta_t) = \frac{\partial }{\partial \Delta_t'} g_\mathrm{free}(\bm{z}_n, \Delta_t') \vert_{\substack{\Delta_t'=\Delta_t}}
\end{equation}
given as
\begin{equation*}
    \bm{J}_{\Delta_t}(\bm{z}_n, \Delta_t) = (\bm{v}_n^\top, (-a_{\rd,n}^2 ||\bm{v}_n|| \bm{v}_n + a_{\rrm,n}^2 (\bm{\omega}_n \times \bm{v}_n) + \bm{g})^\top, \bm{0}, 0, 0)^\top.
\end{equation*}
With Eq.~\ref{eq:jac_t}, $(J3)$ is given by $\bm{J}_{\Delta_t}(\bm{z}^+, \Delta_\mathrm{rem})$.
Next, we decompose $J2$:
\begin{equation*}
    \frac{\partial \bm{z}^+}{\partial \bm{z}_n} = \underbrace{\frac{\partial h(\bm{z}^-)}{\partial \bm{z}^-}}_{J2.1} \underbrace{\frac{\partial \bm{z}^-}{\partial \bm{z}_n}}_{J2.2}
\end{equation*}
As $h(\bm{z}^-) = \bm{C}' \bm{z}^-$ (see Eq. 5 in the main paper), it follows that $(J2.1) = \bm{C}'$. For $(J2.2)$ we have to take into account that $g_\mathrm{free}(\bm{z}_{n}, \Delta_{\mathrm{imp}}(\bm{z}_{n}))$ is a function of the state $\bm{z}_n$ and of the impact time, which again is a function of $\bm{z}_n$:
\begin{equation*}
    \frac{\partial  \bm{z}^-}{\partial  \bm{z}_n} =  \frac{\mathrm{d} g_\mathrm{free}(\bm{z}_n, \Delta_\mathrm{imp}(\bm{z}_n))}{\mathrm{d} \bm{z}_n} = \underbrace{\frac{\partial g_\mathrm{free}(\bm{z}_n, \Delta_\mathrm{imp})}{\partial \bm{z}_n}}_{J2.2.1} +
    \underbrace{\frac{\partial g_\mathrm{free}(\bm{z}_n, \Delta_\mathrm{imp})}{\partial \Delta_\mathrm{imp}}}_{J2.2.2} \underbrace{\frac{\partial \Delta_\mathrm{imp}}{\partial \bm{z}_n}}_{J2.2.3}
\end{equation*}
The terms $(J2.2.1)$, $(J2.2.2)$ are given by the Jacobians $\bm{J}_{\bm{z}}(\bm{z}_n, \Delta_\mathrm{imp})$ and $\bm{J}_{\Delta_t}(\bm{z}_n, \Delta_\mathrm{imp})$, respectively.
For $(J2.2.3)$ let us recapitulate the computation of the impact time
\begin{equation*}
    \Delta_{\mathrm{imp}} = -\left(v_{\rz,n} + \sqrt{v_{\rz,n}^2 + 2 g_{\rz} h}\right)/g_{\rz}
\end{equation*}
with $h = -((p_{\rz,n} - r) - \rz_\mathrm{table})$, i.e.,
\begin{align*}
    \Delta_{\mathrm{imp}} &= -\left(v_{\rz,n} + \sqrt{v_{\rz,n}^2 - 2 g_{\rz} ((p_{\rz,n} - r) - \rz_\mathrm{table})}\right)/g_{\rz} \\
    &= -\left(v_{\rz,n} + \sqrt{v_{\rz,n}^2 - 2 g_{\rz} p_{\rz,n} + 2 g_{\rz} r + 2 g_{\rz} \rz_\mathrm{table} }\right)/g_{\rz}.
\end{align*}
For the derivative w.r.t $\bm{z}_n$ $(J2.2.3)$ it follows that
\begin{equation*}
    \frac{\partial \Delta_\mathrm{imp}}{\partial \bm{z}_n} = [0, 0, \frac{\partial \Delta_\mathrm{imp}}{\partial p_{\rz,n}}, 0, 0, \frac{\partial \Delta_\mathrm{imp}}{\partial v_{\rz,n}}, 0, 0, 0, 0, 0 ]
\end{equation*}
with
\begin{equation*}
    \frac{\partial \Delta_\mathrm{imp}}{\partial p_{\rz,n}} = \frac{1}{\sqrt{v_{\rz,n}^2 - 2 g_{\rz} p_{\rz,n} + 2 g_{\rz} r + 2 g_{\rz} \rz_\mathrm{table} }},
\end{equation*}
\begin{equation*}
    \frac{\partial \Delta_\mathrm{imp}}{\partial v_{\rz,n}} = \frac{-1}{g_\rz} \left( 1 + \frac{v_{\rz,n}}{\sqrt{v_{\rz,n}^2 - 2 g_{\rz} p_{\rz,n} + 2 g_{\rz} r + 2 g_{\rz} \rz_\mathrm{table} }} \right).
\end{equation*}
From $\Delta_\mathrm{rem} = \Delta_T - \Delta_\mathrm{imp}$, it follows that $\frac{\partial \Delta_\mathrm{rem}}{\partial \bm{z}_n} = -\frac{\partial \Delta_\mathrm{imp}}{\partial \bm{z}_n}$, which finally gives $(J4)$.

\bibliography{literature}